\documentclass[runningheads]{llncs}
\usepackage{graphicx}
\usepackage{tikz}
\usepackage{comment}
\usepackage{amsmath,amssymb} % define this before the line numbering.
\usepackage{color}
\usepackage{xspace}
\usepackage{cite}
\usepackage[strings]{underscore} % solve DIO issue
\usepackage{booktabs}
\usepackage{subcaption}

\makeatletter
\DeclareRobustCommand\onedot{\futurelet\@let@token\@onedot}
\def\@onedot{\ifx\@let@token.\else.\null\fi\xspace}

\def\eg{\emph{e.g}\onedot} 
\def\ie{\emph{i.e}\onedot} 
 
 \def\vs{\emph{vs}\onedot}
 
\def\etal{\emph{et al}\onedot}
\makeatother

\begin{document}
% \linenumbers
\pagestyle{headings}
\mainmatter
\def\ECCVSubNumber{3681}  % Insert your submission number here

%*****************
%*****************

\title{A Closer Look at Invariances in Self-supervised Pre-training for 3D Vision} % Replace with your title

\begin{comment}
\titlerunning{ECCV-22 submission ID \ECCVSubNumber} 
\authorrunning{ECCV-22 submission ID \ECCVSubNumber} 
\author{Anonymous ECCV submission}
\institute{Paper ID \ECCVSubNumber}
\end{comment}
%******************

%\begin{comment}
\titlerunning{A Closer Look at Invariances in Self-supervised Pre-training for 3D Vision}
\author{Lanxiao Li \orcidID{0000-0003-3267-2525} \and
Michael Heizmann \orcidID{0000-0001-9339-2055}
}
\authorrunning{L. Li and M. Heizmann}
\institute{Institute of Industrial Information Technology, Karlsruhe Institute of Technology, Karlsruhe, Germany\\
\email{\{lanxiao.li, michael.heizmann\}@kit.edu}}
%\end{comment}

%******************
\maketitle

\begin{abstract}
Self-supervised pre-training for 3D vision has drawn increasing research interest in recent years. In order to learn informative representations, a lot of previous works exploit invariances of 3D features, \eg, perspective-invariance between views of the same scene, modality-invariance between depth and RGB images, format-invariance between point clouds and voxels. Although they have achieved promising results, previous researches lack a systematic and fair comparison of these invariances. To address this issue, our work, for the first time, introduces a unified framework, under which various pre-training methods can be investigated. We conduct extensive experiments and provide a closer look at the contributions of different invariances in 3D pre-training. 
Also, we propose a simple but effective method that jointly pre-trains a 3D encoder and a depth map encoder using contrastive learning. Models pre-trained with our method gain significant performance boost in downstream tasks. For instance, a pre-trained VoteNet outperforms previous methods on SUN RGB-D and ScanNet object detection benchmarks with a clear margin. 

\end{abstract}

% ---------------------------------------------------------
\section{Introduction}
\label{sec:intro}

In order to cope with challenging tasks \eg, object detection, scene understanding, and large-scale semantic segmentation, neural networks for 3D vision are continuously becoming deeper, more complicated, and thus, more data-hungry. 
%However, capturing and annotating data are time-consuming and expensive. 
In recent years, self-supervised pre-training has shown promising progress in natural language processing and computer vision. By learning powerful representations on non-annotated data, the models gain better performance and convergence in downstream tasks. Self-supervised pre-training is especially appealing in 3D vision because 3D annotation is more costly than the 2D counterpart. 

\begin{figure}[t]
\begin{center}
\includegraphics[width=\linewidth]{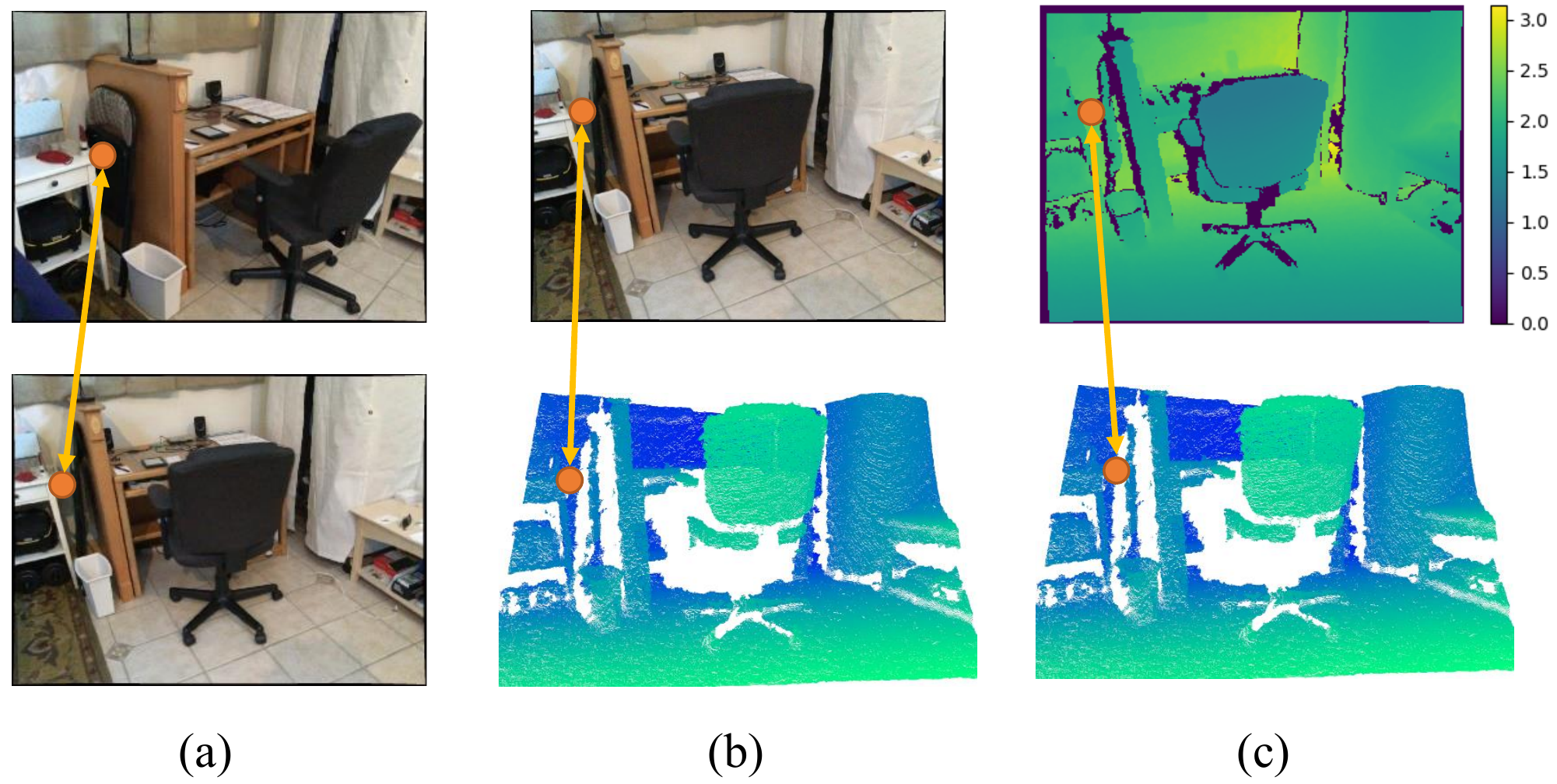}
\end{center}
\caption{
Invariances in contrastive learning for 3D vision. Without loss of generality, we only consider the local correspondence here. Each column includes two views of the same scene. The exemplary correspondences across two views are illustrated with arrows, which means the two points/pixels have the same coordinate in the 3D space. In self-supervised pre-training, the similarity between corresponding local features is maximized, which forces networks to learn invariance between views. \textbf{(a) Perspective-invariance} in two views of the same scene from different view angles. We visualize RGD images instead of point clouds for better clarity. \textbf{(b) Modality-invariance} in an aligned image-point cloud pair. The data formats are also different in this case. But we still refer to it as modality-invariance to distinguish it from the format-invariance within a single modality. \textbf{(c) Format-invariance} between a depth map and a point cloud converted from it. 
}
\label{fig:teaser}
\end{figure}

Self-supervised pre-training for 3D vision has already gained some research interests. A lot of previous works use contrastive learning as a pretext task to pre-train models, as it has shown superior performance in other domains~\cite{PointContrast_eccv_2020, learning_from_2d, DepthContrast, exploring_efficient}. One classic hypothesis in contrastive learning is that a powerful representation should model view-invariant factors. A common approach to creating different views is data augmentation. Moreover, a 3D scene can be captured from various view angles, with different sensors (\eg, RGB and depth cameras) and represented with different formats (\eg, voxels, point clouds, and depth maps)\footnote{To avoid ambiguity, we use the term \emph{data format} instead of \emph{data representation} in this work.}, whereas the major semantic information in the scene is not changed by these factors.
Thus, previous works exploit the perspective-~\cite{PointContrast_eccv_2020, exploring_efficient} , modality-~\cite{learning_from_2d} and format-invariance~\cite{DepthContrast} of 3D features in self-supervised learning, as shown in Fig.~\ref{fig:teaser}. Although these works have shown impressive results, the contribution of the invariances is still under-explored, and a fair and systematic comparison of them hasn't been performed yet. 

In this work, we first establish a unified framework for 3D self-supervised learning. Our framework takes into account the local point/pixel-level correspondence as well as the global instance-level correspondence. Also, our framework unifies contrastive learning with different input data formats and network structures, including Depth-Depth, Point-Point, Depth-Point, Image-Point, and Point-Voxel. By comparing various training strategies exploiting different invariances, we gain non-trivial results. The first insight of this work is that jointly pre-training a 3D encoder and a 2D encoder (Image-Point, Depth-Point) brings better performance than pre-training them separately or jointly pre-training two encoders with the same dimension (\eg, a voxel and a point cloud encoder, which are both three dimensional). 

Also, we propose the simple but effective idea to exploit the format-invariances between depth maps and point clouds/voxels. Our intuition is that depth maps are complementary to point clouds and voxels, although they contain almost the same information. The depth map format has the advantage that it's the natural view of the scene and clearly shows the perspective relationship between objects. Also, real-world depth maps usually contain bad pixels, which means the depth values are unmeasurable. In depth maps, the outlines of unmeasurable regions are sharp and clear, \eg, the chair leg in Fig.~\ref{fig:teaser}~(c). On the contrary, this information is lost if depth maps are lifted into 3D space. Moreover, thanks to its efficiency, the 2D encoder allows high-resolution depth maps as input, which preserves more fine-grained details in data. However, point or voxel-based networks usually take down-sampled or quantized input to avoid the excessive computational cost and memory usage, which results in inevitable information loss. On the other hand, point clouds and voxels are 3D formats and the corresponding networks can directly capture accurate 3D geometry, whereas depth map-based networks learn spatial relationships indirectly. Also, depth maps alone don't contain the information of camera calibrations. By contrasting the features extracted from two complementary data formats, the two networks learn appreciated properties from each other. This simple idea has less requirements on pre-training data and outperforms previous methods in our experiments. 

% The second insight of this work is that the joint pre-training of a 3D encoder with a depth map based encoder generates on pair performance as with a RGB encoder (\ie, a common CNN).

% The single-view 3D data has at least three formats \ie, depth maps, point clouds and voxels. Depth maps are captured directly by 3D sensors \eg, RGB-D cameras. With known camera calibration, depth maps can be lifted into 3D space and converted to point clouds, which can further be quantized to create voxels. Due to different data properties, these data formats are processed by different networks, \eg, 2D CNN for depth maps, PointNet++~\cite{pointnet++} for point clouds and sparse 3D CNN for voxels~\cite{SparseConv_Graham_2018_CVPR}\cite {mink_Choy_2019_CVPR}. 

%Despite minor information loss due to down-sampling, quantization \etc, the major information within a 3D scene is invariant to the data formats (\eg, depth maps, point clouds, voxels). 

% Ideas can be extended to patch-ViT and image-CNN. There are complementary because transformers extract features globally and CNNs extract features locally. Too costly to validate for us.

The contribution of this work is many-fold:
\noindent
\begin{enumerate}
\item We introduce a unified self-supervised pre-training framework for \emph{all} major network types and data formats in 3D vision.
\item We provide a closer look at invariances in 3D pre-training, \eg, format-, perspective- and modality-invariance.
\item We propose a novel approach for 3D self-supervised pre-training, which is based on the format-invariance between depth maps and points/voxels.
\item Our method reaches new SOTA results in multiple downstream tasks \eg, object detection on SUN RGB-D dataset~\cite{sunrgbd} and ScanNet~\cite{dai2017scannet} dataset. 
\item The proposed method is also the first self-supervised pre-training approach for depth map-based networks.
\end{enumerate}

% ---------------------------------------------------------

\section{Related Works}
\label{sec:related}

\noindent
\textbf{Feature Learning with 3D Data.} PointNet~\cite{pointenet_Qi_2017_CVPR} is the pioneer in deep learning methods for point clouds. To aggregate local information, PointNet++~\cite{pointnet++} down-samples and groups point clouds hierarchically. Recent works~\cite{PointCNN_NEURIPS2018, ParamConv_Wang_2018_CVPR, PointConv_Wu_2019_CVPR} define point convolution on point clouds. Voxel-based methods convert irregular point clouds to regular 3D grids and apply 3D convolution~\cite{voxnet15, voxelnet_Zhou_2018_CVPR} or deep sliding windows~\cite{dss_Song_2016_CVPR}. Also, some works~\cite{mink_Choy_2019_CVPR, SparseConv_Graham_2018_CVPR} introduce sparse CNNs to reduce the computational cost and memory footprint. Some other works use 2D CNNs to extract features from depth maps~\cite{25DConv_2019_ICIP, 25Dvotenet, Malleable_25D_Convolution}, LiDAR range images~\cite{RangeRCNN2020, RCDConv2020, FCN4LIDAR_LI_2016} or pseudo images~\cite{PointPillars_Lang_2019_CVPR}. Also, a lot of works use more than one format of 3D data~\cite{pvrcnn_Shi_2020_CVPR, RangeRCNN2020, Point_Voxel_CNN_NEURIPS2019, PointPillars_Lang_2019_CVPR, 3DMV_2018_ECCV}. They and our work share the same motivation to combine the advantages of different data formats. However, our method learns the appreciated property via contrastive learning in the pretext task. In fine-tuning for downstream tasks, only one format is used. 

%\TODO{colorizing, rotation, puzzle, contrastive learning, moco, BYOL, ..}
\noindent
\textbf{Self-supervised Pre-training in Computer Vision.} 
%The goal of self-supervised pre-training is to learn informative features, which can be transferred into downstream tasks by fine-tuning. 
A lot of pretext tasks for self-supervised learning has been proposed. Some generative approaches recover images under some corruption, \eg, auto-encoders for colorization~\cite{zhang_colorful_2016, split_brain_Zhang_CVPR} and denoising~\cite{SSL_denoise_vincent_2008}. Some discriminative approaches generate pseudo-labels for \eg, rotation prediction~\cite{SSL_rotation}, Jigsaw puzzle solving~\cite{noroozi_jagsaw_2016} and objects tracking~\cite{SSL_video_Wang_2015_ICCV}. Recently, contrastive learning achieved impressive performance in self-supervised learning\cite{MOCO_he, Contrastive_Predictive_Coding, max_mutual_info, Misra_2020_CVPR, big_SSmodel, mocov2}. Besides the instance-level discrimination, some works also exploit the local correspondence for better transfer in tasks which need dense features, \eg, object detection and semantic segmentation~\cite{dense_contrast_2021_CVPR, CL_for_global_local, USL_dense_rep}. 

\noindent
\textbf{Self-supervised Pre-training for 3D Data.} Some works~\cite{panos_icml_2018, hassani_iccv_2019, info3d_eccv_2020, sauder_neurips_2019} perform self-supervised learning on synthetic data \eg, ShapeNet~\cite{3D_ShapeNet_2015_CVPR}. However, these approaches don't transfer well to real-world data~\cite{PointContrast_eccv_2020}. PointContrast~\cite{PointContrast_eccv_2020} first uses real-world point cloud data for self-supervised training. It learns perspective-invariance by predicting point-wise correspondence between two partially overlapping point clouds. Liu \etal~\cite{learning_from_2d} pre-train a 3D encoder by using a pre-trained 2D encoder as teacher. On the other hand, Liu \etal~\cite{3Dto2D_distl_Liu_2021_CVPR} propose a distillation pipeline to improve 2D encoders by using geometry guidance from 3D encoders. DepthContrast~\cite{DepthContrast} extends the successful MoCo~\cite{MOCO_he, mocov2} pipeline to 3D domain and exploits the cross-format contrast between point clouds and voxels. Hou \etal~\cite{exploring_efficient} propose spatial partition to improve the contrastive learning and investigate the data-efficiency and label-efficiency of pre-trained models. 

\noindent
\textbf{Multi-modal Feature Fusion.}
The idea of learning from two complementary sources is similar to data fusion. 
In 3D computer vision, a common practice is to fuse the color and geometry information. A lot of fusing approaches have been proposed, \eg, for object detection~\cite{fpointnet_Qi_2018_CVPR, 3dsis_Hou_2019_CVPR, imvotenet_Qi_2020_CVPR, pointfusion_Xu_2018_CVPR} and salient object detection~\cite{25DConv_2019_ICIP, BBS_Net_Fan_ECCV2020, Contrast_Prior_Zhao_2019_CVPR, CMW_li_ECCV2020}.
Some other works use self-supervised pre-training to improve the feature fusion~\cite{P4Contrast, tupleinfonce}. The difference between fusing and contrasting multi-modal features is that fusion enriches features by combining complementary information from different modalities, while contrastive learning maximizes the shared information between modalities. 

% -------------------------------------------------

\section{Method}
In this work, we intend to research the invariances in 3D self-supervised learning, including perspective-, modality- and format-invariance. For a fair comparison, it's meaningful to investigate them under a unified framework. In this section, we first briefly revisit some representative works. Then, we introduce a unified framework to which all previous methods fit. Moreover, we introduce several contrastive learning methods under the unified framework. At last, we provide technical details of the framework. 

\subsection{Unified Framework for 3D Constrastive Learning}
\label{subsec:framework}
In this work, we pay attention to three previous works. (1) PointContrast~\cite{PointContrast_eccv_2020}: it generates two views of the same scene from different perspectives and learns the local correspondence between 3D points using a contrastive loss. 
(2) DepthContrast~\cite{DepthContrast}: following the successful MoCo pipeline~\cite{MOCO_he, mocov2}, it augments two views of the same point cloud to build positive pairs and learn global correspondences by distinguishing the positive samples from a large number of negative samples. Also, it proposes to exploit cross-format contrast between point clouds and voxels.
(3) Pixel-to-point~\cite{learning_from_2d}: its overall pipeline is similar to PointContrast. However, it learns local correspondences between point clouds and RGB images, in order to benefit from strong pre-trained RGB encoders.

Therefore, a unified framework must support both the local and global correspondence of 3D data and at least two different input types, either from different modalities (\eg, RGB images and point clouds) or with different data formats (\eg, depth maps and point clouds). 

\begin{figure}
    \centering
    \includegraphics[width=\linewidth]{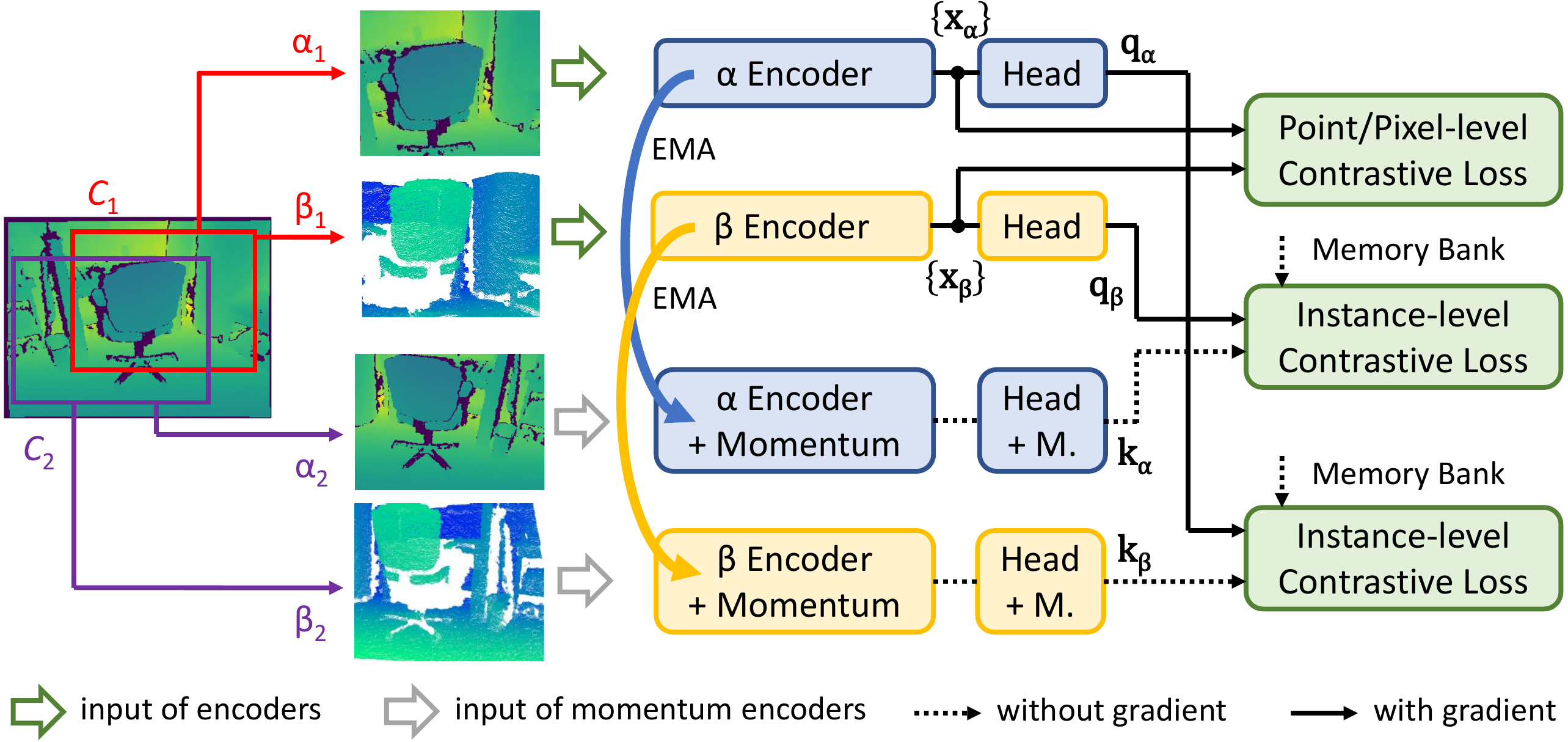}
    \caption{A unified framework for 3D contrastive learning. Here, $\alpha$ and $\beta$ refer to data types, \eg, point clouds, images, and depth maps. }
    \label{fig:pipeline}
\end{figure}

We show our framework in Fig.~\ref{fig:pipeline}, which uses a single-view depth map or RGB-D image for simplicity. However, experiments show that the pre-trained weights generalize well on reconstructed multi-view 3D scans. Without loss of generality, we assume the input of the framework is a depth map in this section. We randomly crop the input to get crop $C_1$, which is further randomly augmented and converted into views $\alpha_1$ and $\beta_1$. Here $\alpha$ and $\beta$ refer to data formats (\eg, depth maps and point clouds, as visualized in Fig.~\ref{fig:pipeline}). Then, $\alpha_1$ and $\beta_1$ go through respective encoders and are encoded into pixelwise or pointwise features $\{\mathbf{x_{\alpha}}\}$ and $ \{ \mathbf{x_{\beta}} \}$. Note that the $\alpha$ and $\beta$ encoders are usually different networks matching the input formats. But in the case of $\alpha=\beta$, they share the weights, following~\cite{PointContrast_eccv_2020}. As $\alpha_1$ and $\beta_1$ are generated from the same crop $C_1$, the dense correspondence between $\{\mathbf{x_{\alpha}}\}$ and $ \{ \mathbf{x_{\beta}} \}$ can be easily calculated without camera extrinsic. In this work, we follow~\cite{PointContrast_eccv_2020} and use InfoNCE loss to train dense local correspondence, which is further explained in Sec.~\ref{subsec:details}. 

In order to learn informative representations, our framework also considers the global correspondence between views. Following \cite{DepthContrast}, we perform instance discrimination based on global features $\mathbf{q_\alpha}$ and $\mathbf{q_\beta}$, which are globally pooled and projected from $\{ \mathbf{x_\alpha} \}$ and $\{ \mathbf{x_\beta} \}$. 
%using respective projection heads consisting of a global pooling and a multi-layer perceptron (MLP). 
To preserve a large number of negative samples for effective contrastive learning, we use memory banks and momentum encoders,
%whose weights are exponentially moving averaged (EMA) from respective encoders,
following the successful MoCo pipeline~\cite{MOCO_he, mocov2}. However, in supplementary material, we further show that our methods can be generalized to other pipelines, \eg BYOL \cite{BYOL} and SimSiam \cite{simsiam_Chen_2021_CVPR}.

Analog to crop $C_1$, we randomly crop $C_2$ from the same depth map, generate $\alpha_2$ and $\beta_2$ and feed them to the momentum encoders. We refer to the globally pooled and projected features from momentum encoders as $\mathbf{k_\alpha}$ and $\mathbf{k_\beta}$, respectively. They are dynamically saved and updated in memory banks during training. Note that unlike~\cite{DepthContrast} our work only contrasts features from different input formats, as we empirically found the gains from additional contrast within the formats are marginal.

\subsection{Variants of Strategies}

\begin{figure}
    \centering
    \includegraphics[width=\linewidth]{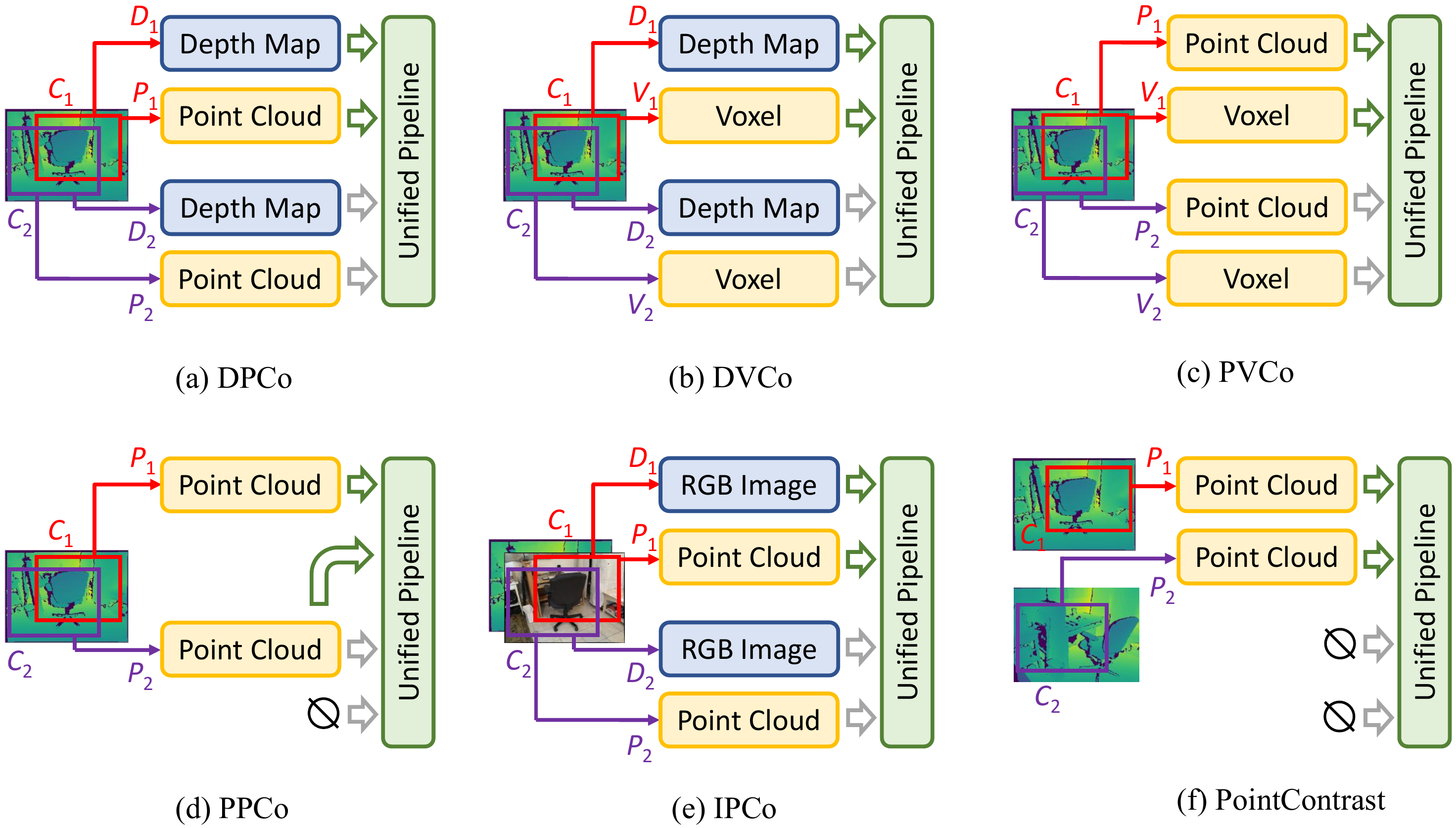}
    \caption{Contrastive learning strategies under a unified framework.}
    \label{fig:variants}
\end{figure}

As the overall framework is shown, we now introduce various contrastive learning strategies under this framework. As shown in Fig.~\ref{fig:variants}, we investigate the following variants in this work:
\begin{enumerate}
    \item DPCo (Depth-Point Contrast), our proposed method, which learns \emph{format-invariance} between depth maps and point clouds.
    \item DVCo (Depth-Voxel Contrast), our proposed method, which learns \emph{format-invariance} between depth maps and voxels.

    \item PVCo (Point-Voxel Contrast), which learns \emph{format-invariance} between point clouds and voxels. It's extended from PointContrast~\cite{DepthContrast}.

    \item PPCo (Point-Point Contrast), which only uses point clouds as input. It serves as a baseline method as it only learns invariance against data augmentation. 
    
    \item IPCo (Image-Point Contrast), which learns \emph{modality-invarince} between RGB images and point clouds. It's inspired by Pixel-to-point~\cite{learning_from_2d}.
    \item PointContrast~\cite{PointContrast_eccv_2020}, which learns \emph{perspective-invariance} between view angles. It can be interpreted as a special case of our unified framework, since it generates the crops $C_1$ and $C_2$ from two overlapping depth maps from different view angles and only considers the local correspondence. 
\end{enumerate}

In this work, we propose to contrast a 3D format and a 2D format of the same geometric data (\ie, DPCo and DVCo). Although they represent the same 3D scene, the two formats are complementary to some extent. As discussed in Sec.~\ref{sec:intro}, point clouds and voxels directly represent 3D geometry while having inevitable information loss due to sampling and bad pixels. On the contrary, depth maps reserve more information but only represent the 3D scene indirectly. Experiment results show that our methods bring significantly better performance than PPCo and PVCo, which contrast only 3D formats.

% TODO: how good is IPCo???

\subsection{Details}
\label{subsec:details}
\noindent
\textbf{Point Cloud Encoder.} We use a U-shaped PointNet++\cite{pointnet++} and follow the network configuration in \cite{votenet_Qi_2019_ICCV}, which consists of 4 down-sampling and 2 up-sampling modules. We use 20K points as input in the pre-training. The number of output points is fixed to 1024. 

\noindent
\textbf{Voxel Encoder.} We use a sparse residual U-Net~\cite{mink_Choy_2019_CVPR} with 34 convolution layers to encode voxel inputs, following previous works~\cite{PointContrast_eccv_2020, exploring_efficient}. We use the implementation of sparse convolution in \cite{mink_Choy_2019_CVPR}. For geometry-only input, we set all input features to 1. For input with colors, we use normalized RGB values as input features. In pre-training, we quantize inputs with the voxel size of 2.5 cm. The output of the voxel encoder has the same resolution as the input. 

\noindent
\textbf{Depth Map Encoder.} We use the U-shaped 2D CNN in~\cite{25Dvotenet} as depth map encoder. The network is a modified ResNet-34~\cite{resnet_He_2016_CVPR} with relative depth convolution~\cite{25Dvotenet} and extra up-sampling layers. The input is resized and zero-padded to 352$\times$352. The output is a feature map down-sampled with factor 8. 

\noindent
\textbf{Color Image Encoder.} Analog to the depth map encoder, we use a ResNet-34 with extra up-sampling layers to encoder the RGB images. We initialize this encoder with the pre-trained weights on ImageNet~\cite{imagenet}, following the setup in~\cite{learning_from_2d}.

\noindent
\textbf{Momentum Encoders and Projection Heads.} The momentum encoders have the same structure as the encoders. Their weights are updated via exponential moving average (EMA) from the corresponding encoders instead of back-propagation. We use global max pooling to aggregate the global features. The pooling layer is followed by an MLP consisting of 3 fully connected layers. The intermediate and the output layer have 512 and 128 channels, respectively. The projection heads of momentum encoders are updated via EMA as well.  

\noindent
\textbf{Loss Functions.} Our loss function consists of a local sub-loss $L_\mathrm{l}$ and a global sub-loss $L_\mathrm{g}$. The local sub-loss is an InfoNCE loss which optimizes the local dense correspondence: 

\begin{equation}
    L_\mathrm{l, \, \alpha\beta} = - \sum_{i} \log \frac{\exp(\mathbf{x}_\mathrm{\alpha, \, i} \cdot \mathbf{x}_\mathrm{\beta, \, i} / \tau)}
    {\exp(\mathbf{x}_\mathrm{\alpha, \, i} \cdot \mathbf{x}_\mathrm{\beta, \, i} / \tau) + \sum_{j \neq i} \exp(\mathbf{x}_\mathrm{\alpha, \, i} \cdot \mathbf{x}_\mathrm{\beta, \, j} / \tau)}
\end{equation}
With $\mathbf{x}_\mathrm{\alpha, \, i} \in \{ \mathbf{x}_\alpha \}$ and $\mathbf{x}_\mathrm{\beta, \, j} \in \{ \mathbf{x}_\beta \}$. If the corresponding 3D coordinates of feature vector $\mathbf{x}_\mathrm{\alpha, \, i}$ and $\mathbf{x}_\mathrm{\beta, \, j}$ are close, they're considered as a positive pair and have $i=j$. The temperature $\tau$ is a hyperparameter and is set to 0.07 in this work. All features are L2-normalized before being fed into the loss function.

The global sub-loss is applied to optimize an instance discrimination task: 

\begin{equation}
    L_\mathrm{g, \, \alpha \beta} = - \log \frac{\exp(\mathbf{q}_\mathrm{\alpha} \cdot \mathbf{k}_\mathrm{\beta} / \tau)}
    {\exp(\mathbf{q}_\mathrm{\alpha} \cdot \mathbf{k}_\mathrm{\beta} / \tau) + \sum_{n=1}^{N-1} \exp(\mathbf{q}_\mathrm{\alpha} \cdot \mathbf{k}_\mathrm{\beta, \, n} / \tau)}
\end{equation}
The vector $\mathbf{q}_\mathrm{\alpha}$ refers to the global feature from the $\alpha$ encoder and $\mathbf{k}_\mathrm{\beta}$ the global feature from the $\beta$ momentum encoder. Since $\mathbf{q}_\mathrm{\alpha}$ and $\mathbf{k}_\mathrm{\beta}$ are generated from the same data sample, they make a positive pair. Features $\mathbf{k}_\mathrm{\beta, \, n}$ correspond to other samples and are read from a memory bank with the size $N$. We use $N=2^{15}$ in this work. Following previous works, we make our loss symmetric to $\alpha$ and $\beta$. The total loss can be formulated as
\begin{equation}
    L = L_\mathrm{l} + L_\mathrm{g} = 0.25 \cdot (L_\mathrm{l, \, \alpha \beta} + L_\mathrm{l, \, \beta \alpha} +  L_\mathrm{g, \, \alpha \beta} + L_\mathrm{g, \, \beta \alpha}) \, . 
\end{equation}
Principally, $L$ can be a weighted sum of $L_\mathrm{l}$ and $L_\mathrm{g}$ and the weighting factors can be tuned. But we empirically found that the simple arithmetic average already generates good results.

\noindent
\textbf{Data Augmentation.} We randomly crop $C_1$ and $C_2$. Also, we randomly drop a square area in each crop. We apply random rotation, scaling, and flipping to the point clouds and voxels. We randomly rotate depth maps around principal points and set 20\% pixels on the depth map to zero. For RGB images, we apply random color jitter, grayscale, and Gaussian blur. 

\noindent
\textbf{Dataset.}
We use ScanNet~\cite{dai2017scannet} for the pre-training, following previous works~\cite{PointContrast_eccv_2020, DepthContrast, exploring_efficient, P4Contrast}. ScanNet is a large-scale indoor dataset, which contains about 1500 scans reconstructed from 2.5M RGB-D frames. We follow the official train/val split and sample 78K frames (one in every 25 frames) from the train set. 

\noindent
\textbf{Training.} We pre-train the encoders for 120 epochs. We use SGD optimizer with momentum of 0.9 and an initial learning rate of 0.03. We train models on two NVIDIA Tesla V100 GPUs with a total of 64 GB memory and use as large batch size as it fits. The batch size of different strategies varies from 32 to 64. The learning rate is decayed with a cosine schedule. The pre-training takes from two to four days using PyTorch with Distributed Data Parallel. 

More technical details can be found in the supplementary material.

%\begin{equation}
%    L_\mathrm{local} = - \sum_{(i,j)\in P_l, \, \mathbf{x}_\mathrm{\alpha, i} \in \{ \mathbf{x}_\alpha \}} \log \frac{\exp(\mathbf{x}_\mathrm{\alpha, i} \cdot \mathbf{x}_\mathrm{\beta, j} / \tau)}
%   {\sum_{\mathbf{x}_\mathrm{\beta, k} \in \{ \mathbf{x}_\beta \}} \exp(\mathbf{x}_\mathrm{\alpha, i} \cdot \mathbf{x}_\mathrm{\beta, k} / \tau)}
%\end{equation}

% ------------------------------------------------------------------

\section{Experiments and Results}
\label{sec:experiments}
In this section, we first briefly introduce the experimental setups. Then, we compare and analyze different contrastive learning strategies in detail under our unified framework, to clarify the contribution of the invariances. Then, we compare our method (DPCo) with state-of-the-art methods in the point cloud object detection task. At last, we show transfer learning results of our methods on voxels and depth maps. More experimental results can be found in the supplementary material.

% We report the performance of pre-training methods on three sets of representative down-stream tasks, categorized according to the input formats. (1) Point transfer: We use point and depth map inputs in pre-training and fine-tune a VoteNet for 3D object detection. (2) Depth transfer. We use point and depth map inputs in pre-training and fine-tune a 2.5D-VoteNet for 3D object detection. Setups for fine-tuning follow the previous works~\cite{votenet_Qi_2019_ICCV, mink_Choy_2019_CVPR, 25Dvotenet}. (3) Voxel transfer: We use voxel and RGB-D inputs in pre-training and fine-tune a Sparse 3D ResNet~\cite{mink_Choy_2019_CVPR} for 3D semantic segmentation. 

\subsection{Invariances in 3D Self-supervised Pre-training}
In this subsection, we focus on the performance of transfer learning on point cloud-based 3D detection task, since we believe the 3D detection reflects the encoder's capability of capturing both semantic (\ie, objects classification) and geometric (\ie, bounding box regression) information and is thus representative. Also, 3D detection using raw points is well studied in previous works~\cite{fpointnet_Qi_2018_CVPR, votenet_Qi_2019_ICCV, imvotenet_Qi_2020_CVPR, HGNet_Chen_2020_CVPR}. In this work, we fine-tune a VoteNet~\cite{votenet_Qi_2019_ICCV} with a PointNet++ backbone on SUN RGB-D~\cite{sunrgbd} and ScanNet~\cite{dai2017scannet} object detection benchmark. The evaluation metrics are the mean Average Precision over the representative classes with the threshold of 25\% and 50\% 3D-IoU (\ie, AP25 and AP50). 

\noindent
\textbf{Comparison under the Unified Framework.}
In this experiment, we compare various contrastive learning strategies under our unified framework. As shown in Tab.~\ref{tab:invariances}, all pre-training methods deliver better results than training from scratch in both 3D detection benchmarks. Note that ScanNet benchmark uses point clouds reconstructed from multiple views. Our unified framework, which assumes that the pre-training data are independent single depth maps or RGB-D images, still significantly improves the detection results on this dataset. It implies that the weights pre-trained on single-view data generalize well on multi-view data. 

The baseline strategy PPCo utilizes solely the invariance against data augmentation. However, it surpasses PointContrast, which relies on extrinsic parameters, in two out of four metrics. It implies that with a proper design (in our case, the local dense contrast and the MoCo-style instance discrimination), the perspective-invariance is unnecessary in pre-training. A similar observation is also reported in~\cite{DepthContrast}. We hypothesize that in the instance discrimination sub-problem, the network has to distinguish inputs from very similar view angles, as we extract training data from continuous RGB-D videos. This can be interpreted as hard example mining, which forces the network to focus on perspective-relevant details. Thus, with the help of global correspondence in pre-training, the encoders implicitly learn perspective-relevant information, but not necessarily the invariance in this case. 

\begin{table}[ht]
    \centering
    \begin{tabular}{l|c|c|cc|cc}
        \toprule
         \textbf{Method} & \textbf{Invariance} & \textbf{Correspond.} & \multicolumn{2}{c|}{\textbf{SUN RGB-D}} & \multicolumn{2}{c}{\textbf{ScanNet}} \\
         \cline{4-7}
         &  &  & \textbf{AP25} & \textbf{AP50} & \textbf{AP25} & \textbf{AP50} \\
         \midrule
         From Scratch & - & - & 58.4 & 33.3 & 60.0 & 37.6 \\
         \midrule
         PPCo & Augmentation & Local+Global & 58.6 & 34.9 & 62.6 & 39.5 \\
         PointContrast & Perspective & Local & 59.6 & 34.1 & 62.8 & 38.1 \\
         PVCo & Format (3D-3D) & Local+Global & 59.3 & 34.9 & 62.8 & 39.5 \\
         IPCo & Modality & Local+Global & \textbf{60.2} & 35.5 & 63.9 & 40.9 \\
         DPCo (Ours) & Format (2D-3D) & Local+Global & 59.8 & \textbf{35.6} & \textbf{64.2} & \textbf{41.5} \\
         \bottomrule
    \end{tabular}
    \caption{VoteNet fine-tuning performance of self-supervised pre-training strategies with different invariances. We reproduce the results without pre-training using the open-source code of~\cite{votenet_Qi_2019_ICCV}, which are slightly better than the original publication. }
    \label{tab:invariances}
\end{table}

Moreover, PVCo, which contrasts features from point clouds and voxels, brings slightly better though very similar results as PPCo. It's probably due to the nature of point clouds and voxels as they both represent 3D coordinates directly. Also, PointNet++ is similar to 3D ConvNets, as it conducts convolution-like local aggregation, uses shared weights in a sliding window manner, and has a hierarchical topology with sub- and up-sampling. Thus, jointly pre-training voxel and point cloud encoders bring limited benefits to the point cloud encoder, compared to pre-training it alone. In this case, incorporating voxel features can be interpreted as a strong data augmentation to point clouds. 

However, IPCo and DPCo, which contrast a 2D data format (\eg, color images or depth maps) and a 3D format (\eg, point clouds) achieve significantly better results than PPCo and PVCo, which utilize only 3D formats. It confirms our intuition that the 2D data format is complementary to the 3D format and the correspondence between them can provide strong contrast in self-supervised pre-training. More interestingly, our proposed method DPCo, which uses solely the geometrical information, reaches on-par or better performance as the one using both geometrical and color inputs (IPCo). This is an important advantage in practice, as our method is applicable even if the RGB images are not available or hard to align with depth maps. It also implies that the performance gains of IPCo come probably not from the color information, but from other factors \eg, different resolutions and perspective fields of 2D and 3D networks. Another advantage of DPCo is that it trains faster than PPCo and PVCo, thanks to the efficiency of 2D CNN. 

\noindent
\textbf{Local and Global Correspondence.} 
Our unified framework supports both the local and global correspondence of 3D data in the pre-training. In the following experiments, we investigate the contribution of each type of correspondence separately. As shown in Tab.~\ref{tab:local_contrast} and Tab.~\ref{tab:global_contrast}, using local and global correspondence alone in the pre-training improves the performance of encoders. 
Also, comparing with Tab.~\ref{tab:invariances}, it's clear that combining them can bring further improvement, which is also observed in 2D pre-training, as discussed in~\cite{dense_contrast_2021_CVPR}. Moreover, Tab.~\ref{tab:local_contrast} and Tab.~\ref{tab:global_contrast} show similar trends as Tab.~\ref{tab:invariances}, where IPCo and DPCo show superior performance over others. Interestingly, in Tab.~\ref{tab:local_contrast} IPCo and DPCo achieve better results than PointContrast even without the global correspondence.

% Note that the two views $\alpha_1$ and $\beta_1$ and generated from exactly the same crop, while the two views in PointContrast are from different view angles. 

\begin{table}[ht]
    % left
    \parbox{.48\textwidth}{
    \centering
    \begin{tabular}{l|cc|cc}
        \toprule
         \textbf{Contrast} & \multicolumn{2}{c|}{\textbf{SUN RGB-D}} & \multicolumn{2}{c}{\textbf{ScanNet}} \\
         \cline{2-5} & \textbf{AP25} & \textbf{AP50} & \textbf{AP25} & \textbf{AP50} \\
         \midrule
         w/o & 58.4 & 33.3 & 60.0 & 37.6 \\
         \midrule
         PPCo & 58.7 & 34.8 & 62.2 & 38.8 \\
         PVCo & 59.1 & 34.6 & 62.2 & 39.0 \\
         PointCo. & 59.6 & 34.1 & 62.8 & 38.1 \\
         IPCo & \textbf{60.1} & \textbf{35.6} & 62.5 & 39.4 \\
         DPCo & 59.6 & 35.1 & \textbf{64.2} & \textbf{40.5} \\
         
         % IPCo & 60.1 & \textbf{35.5} & \textbf{61.3} & \textbf{38.0} \\
         % DPCo & \textbf{60.2} & \textbf{35.5} & 61.0 & \textbf{38.0} \\
         % NOTE: with Adam optimizer, worse results
         \bottomrule
    \end{tabular}
    \caption{Different choices of local correspondence in pre-training.}
    \label{tab:local_contrast}
    }
    \hfill
    % right
\parbox{.48\textwidth}{
    \centering
    \begin{tabular}{l|cc|cc}
        \toprule
         \textbf{Contrast} & \multicolumn{2}{c|}{\textbf{SUN RGB-D}} & \multicolumn{2}{c}{\textbf{ScanNet}} \\
         \cline{2-5} & \textbf{AP25} & \textbf{AP50} &  \textbf{AP25} & \textbf{AP50} \\
         \midrule
         w/o & 58.4 & 33.3 & 60.0 & 37.6 \\
         \midrule
         PPCo & 59.3 & 35.1 & 62.7 & 39.3 \\
         PVCo & 59.0 & \textbf{35.3} & 62.5 & 39.6 \\
         IPCo & \textbf{59.4} & 34.5 & 63.3 & 40.2 \\
         DPCo & \textbf{59.4} & 34.9 & \textbf{63.8} & \textbf{41.0} \\
         \bottomrule
    \end{tabular}
    \caption{Different choices of global correspondence in pre-training.}
    \label{tab:global_contrast}
    }
\end{table}

\noindent
\textbf{Summary. } Our observations concerning with the invariances can be summarized as follows:
\begin{enumerate}
    \item Explicit perspective-invariance in 3D self-supervised learning is unnecessary. 
    \item Format-invariance between 3D formats (\eg, point clouds and voxels) improves the performance but the gains are marginal. 
    \item Format-invariance between depth map and a 3D formats (\eg, depth maps and point clouds) significantly improves the performance, which is slightly better than modality-invariance between point clouds and RGB-images but has fewer requirements on the training data. 
\end{enumerate}

\subsection{Comparison with SOTA Methods} 

In the previous subsection, our proposed method DPCo shows the best performance among all variants. In this subsection, we compare it with other SOTA self-supervised pre-training methods. Still, we use the fine-tuning performance in point cloud object detection task as the metric. To obtain a strong supervised baseline, we follow the setup in~\cite{25Dvotenet} and generate bounding box annotations for single frames in ScanNet. Then, we pre-train a VoteNet with full supervision. For a fair comparison, the supervised baseline and other self-supervised methods use exactly the same number of frames for pre-training. 

In Tab.~\ref{tab:det_fintune}, we compare our methods with PointContrast~\cite{PointContrast_eccv_2020}, DepthContrast~\cite{DepthContrast}, pixel-to-point~\cite{learning_from_2d} and the method of Hou \etal~\cite{exploring_efficient}, which are already discussed in Sec.~\ref{sec:related} and Sec.~\ref{subsec:framework}. As Tab.~\ref{tab:det_fintune} shows, our method outperforms other self-supervised pipelines in three metrics out of four. It even outperforms the fully supervised baseline on ScanNet AP25 and AP50. Also, our method has on-par performance on SUN RGB-D AP50 und ScanNet AP25 with the up-scaled version of DepthContrast~\cite{DepthContrast}, which uses a 3-times larger network and is pre-trained with 5-times more data. This result implies that the contribution of format-invariance between point clouds and depth maps is comparable with scaling up the model capacity and the data amount. Also note that besides depth maps (or the equivalence \eg, range images) and camera intrinsic, which are available in almost all 3D datasets, our method doesn't require any extra data, \eg, color images and camera extrinsic, while a lot of SOTA methods do~\cite{PointContrast_eccv_2020, exploring_efficient, learning_from_2d}. 

\begin{table}[ht]
    \centering
    \begin{tabular}{l|cc|cc}
        \toprule
         \textbf{Pre-training} & \multicolumn{2}{c|}{\textbf{SUN RGB-D}} & \multicolumn{2}{c}{ \textbf{ScanNet}} \\
         \cline{2-5} & \textbf{AP25} & \textbf{AP50} & \textbf{AP25} & \textbf{AP50} \\
         \midrule
         From Scatch & 58.4 & 33.3 & 60.0 & 37.6 \\
         \midrule
         PointContrast~\cite{PointContrast_eccv_2020} & - & 34.8 & - & 38.0 \\
         PointContrast (Ours) & 59.5 & 34.0 & 61.6 & 38.2 \\
         Hou~\etal~\cite{exploring_efficient} & - & - & - & 39.2 \\
         Pixel-to-point~\cite{learning_from_2d} & 57.2 & 33.9 & 59.7 & 38.9 \\
         Pixel-to-point~(Ours) & 60.1 & \textbf{35.6} & 62.5 & 39.4 \\
         DepthContrast~\cite{DepthContrast} & \textbf{60.4} & - & 61.3 & - \\
         DPCo (Ours) & 59.8 & \textbf{35.6} & \textbf{64.2} & \textbf{41.5} \\
         \midrule
         \textcolor{gray}{DepthContrast\dag~\cite{DepthContrast}} & 61.6 & 35.5 & 64.0 & 42.9 \\
         \textcolor{gray}{Supervised} & 62.0 & 36.3 & 61.9 & 38.6 \\
         \bottomrule
    \end{tabular}
    \caption{Fine-tuning results of VoteNet on SUN RGB-D and ScanNet (scan-level) object detection benchmark with different pre-traning methods. The absent values are not reported in original publications. We report the results of PointContrast and pixel-to-point with our own implementations, as the original papers use a voxel-based backbone instead of a PointNet++. \textcolor{gray}{Grayed methods} refer to results with extra data or annotations. Specifically, DepthContrast\dag~\cite{DepthContrast} uses a scaled PointNet++ backbone with 3$\times$ more parameters and is pre-trained on both ScanNet and Redwood indoor RGB-D scan dataset~\cite{Park2017}.}
    \label{tab:det_fintune}
\end{table}

\subsection{Data Efficiency}
\begin{figure}[h]
\parbox{.48\textwidth}{
    \centering
    \includegraphics[height=44mm]{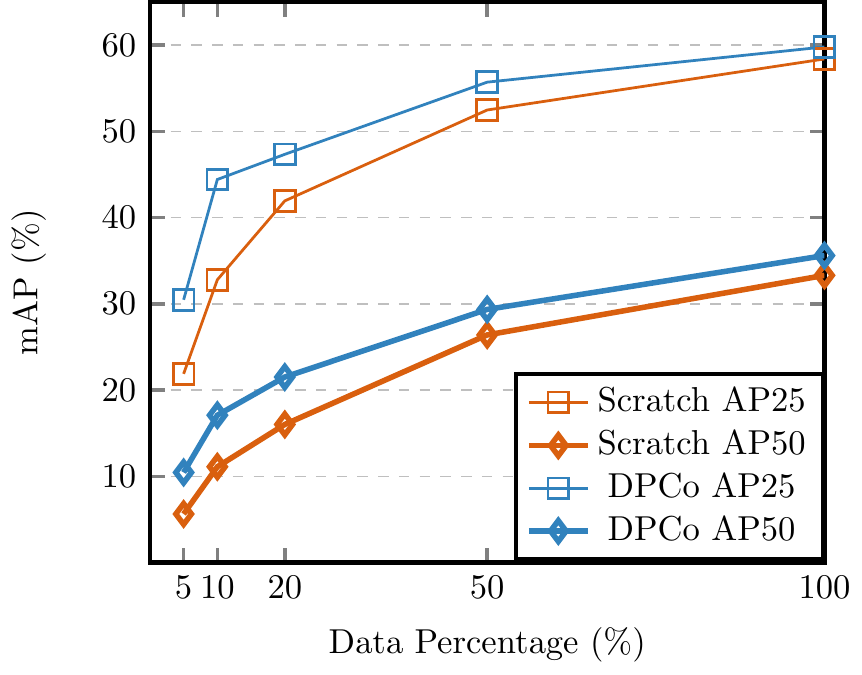}
    \caption{Detection results on SUN RGB-D with reduced amount of data.}
    \label{fig:data_eff_sun}
}
\hfill
\parbox{.48\textwidth}{
    \centering
    \includegraphics[height=44mm]{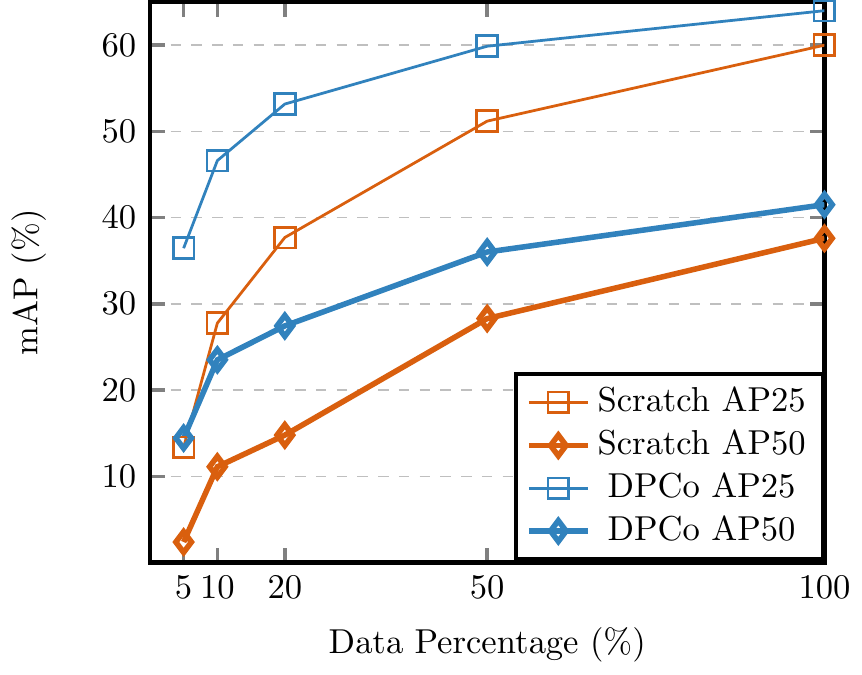}
    \caption{Detection results on ScanNet with reduced amount of data. }
    \label{fig:data_eff_scannet}
}
\end{figure}

One important goal of pre-training is to transfer the features to very small datasets. To simulate this scenario, we randomly sample a small partition from the downstream datasets (\eg, 5\%, 10\%) and fine-tune a VoteNet with the backbone pre-trained by DPCo. Experiments with the same percentage share the same data samples. The validation set is not sampled.
As shown in Fig.~\ref{fig:data_eff_sun} and Fig.~\ref{fig:data_eff_scannet}, the pre-training brings more improvement, when less fine-tuning data are available. The trend is more obvious on ScanNet, as it contains fewer training samples than SUN RGB-D (1.2K \vs 5K in total). Especially, the DPCo pre-training boosts the AP25 on ScanNet from 13.3\% to 36.5\% and the AP50 from 2.4\% to 14.4\%, when only 5\% of training data are used.

\subsection{Transfer on Depth Map and Voxel Encoders}
Till now, we only showed the transfer learning results of point cloud encoders (PointNet++). In this subsection, we investigate the generalization of our methods (DPCo and DVCo) on the depth map and voxel encoders. 

\noindent
\textbf{Depth Map Encoders.} We fine-tune a 2.5D-VoteNet~\cite{25Dvotenet}, which is a variant of VoteNet with a depth map-based backbone, by initializing its backbone with the pre-trained weights. In order to clarify the contribution of format-invariance, we also pre-train the depth map encoder with solely depth map input. This strategy is similar to PPCo in~Fig.~\ref{fig:variants} and we name it DDCo (Depth-Depth Contrast). Since 2.5D-VoteNet doesn't support multi-view input, we only fine-tune it on SUN RGB-D dataset. One surprising result in Tab.~\ref{tab:depth} is that the pre-training using DDCo degrades the performance. As a depth map is an indirect representation of 3D coordinates, we hypothesize that DDCo makes the depth map encoder focus on the 2D textures instead of the true 3D geometry, which can be interpreted as cheating in the pre-training. It also implies that the pre-training of depth map encoders is non-trivial and requires a careful design. 
However, our proposed methods DPCo and DVCo consistently improve the detection results. Since the point cloud and voxel encoders are able to capture 3D geometrical information by their nature, they can provide guidance to the depth map encoder and prevent the depth map encoder from paying too much attention to 2D patterns. Also, combined with the results in Tab.~\ref{tab:det_fintune}, it's worth noticing that DPCo improves the 3D and 2D encoders at the same time. It proves that the principle of our methods is different from knowledge distillation (KD), which uses a stronger model as a teacher to improve a weaker student model.

\begin{table}[ht]
\parbox{.48\textwidth}{
    \centering
    \begin{tabular}{l|c|c}
         \toprule
         \textbf{Pre-training} & \textbf{AP25} & \textbf{AP50} \\
         \midrule
         From Scratch & 60.8 & 36.9 \\
         \midrule
         DDCo & 56.0 & 31.2 \\
         DVCo & 61.0 & \textbf{39.3} \\
         DPCo & \textbf{61.4} & 38.8 \\
         \bottomrule
    \end{tabular}
    \caption{Fine-tuning results of 2.5D-VoteNet on SUN RGB-D dataset with different contrasting strategies. }
    \label{tab:depth}
}
\hfill
\parbox{.48\textwidth}{
    \centering
    \begin{tabular}{l|c|c}
         \toprule
         \textbf{Pre-training} & \textbf{S3DIS} & \textbf{ScanNet} \\
         \midrule
         From Scratch & 66.1 & 69.6 \\
         \midrule
         PVCo & 66.6 & 70.3 \\
         DVCo & \textbf{67.2} & \textbf{70.5} \\
         \bottomrule
    \end{tabular}
    \caption{Fine-tuning results of Sparse 3D ResNet in semantic segmentation tasks. The evaluation metric is mean IoU over classes (mIoU).}
    \label{tab:voxel}

}
\end{table}

\noindent
\textbf{Voxel Transfer.} To evaluate our methods on voxel-based networks, we use DVCo to pre-train a voxel encoder and fine-tune it for semantic segmentation on ScanNet~\cite{dai2017scannet} and S3DIS~\cite{s3dis_dataset_17} dataset. We compare the performance with the not pre-trained baseline and PVCo. As shown in Tab.~\ref{tab:voxel}, DVCo brings significant improvement to the baseline on both segmentation tasks. The performance is also higher than PVCo, which is consistent with the transfer learning results of point cloud encoders. 

%\subsection{Ablation Studies}
%\noindent
%\textbf{Constrastive Schemes.}

%\noindent
%\textbf{Convergence and Optimizers. }

\section{Conclusion and Future Works}
In this work, we establish a unified framework to fairly compare the contribution of perspective-, format- and modality-invariance in 3D self-supervised learning. With the help of our framework, we find out that contrasting a 3D data format (\eg point clouds and voxels) with a 2D data format (\eg images and depth maps) is especially beneficial. Moreover, we propose to contrast point clouds or voxels with depth maps instead of RGB images, which brings better performance and has fewer requirements on the training data than previous methods. Experimental results show that our methods improve all types of encoders in 3D vision, including point cloud, voxel, and depth map encoders. 

Furthermore, some concerns deserve more research effort. For instance, in our framework, we jointly pre-train two different encoders. Although they both gain performance boost in downstream tasks, it's still unclear, whether each encoder has reached the optimum in the pre-training. In future work, we intend to investigate the optimization and convergence of the joint pre-training.

\noindent
\textbf{Acknowledgement.} Parts of this work were financed by Baden-Württemberg Stiftung gGmbH within the project KOMO3D.

\clearpage
\bibliographystyle{splncs04}
\bibliography{egbib}
\end{document}